\documentclass{article}

\PassOptionsToPackage{numbers, compress}{natbib}

\usepackage[preprint]{neurips_data_2023}

\usepackage[utf8]{inputenc} 
\usepackage[T1]{fontenc}    
\usepackage{hyperref}       
\usepackage{url}            
\usepackage{booktabs}       
\usepackage{amsfonts}       
\usepackage{nicefrac}       
\usepackage{microtype}      
\usepackage{xcolor}         
\usepackage{multirow,graphicx,bm,caption,subfig,tabularx}

\title{Revisiting pre-trained remote sensing model benchmarks: resizing and normalization matters}

\author{
Isaac Corley\thanks{Equal contribution}\\
University of Texas at San Antonio\\
San Antonio, TX, USA\\
\texttt{isaac.corley@my.utsa.edu} \\
\And
Caleb Robinson\footnotemark[1]\\
Microsoft AI for Good Research Lab\\
Redmond, WA, USA\\
\texttt{caleb.robinson@microsoft.org} \\
\And
Rahul Dodhia\\
Microsoft AI for Good Research Lab\\
Redmond, WA, USA\\
\texttt{rahul.dodhia@microsoft.com} \\
\And
Juan M. Lavista Ferres\\
Microsoft AI for Good Research Lab\\
Redmond, WA, USA\\
\texttt{jlavista@microsoft.com} \\
\And
Peyman Najafirad\\
University of Texas at San Antonio\\
San Antonio, TX, USA\\
\texttt{peyman.najafirad@utsa.edu} \\
}

\begin{document}

\maketitle

\begin{abstract}
Research in self-supervised learning (SSL) with natural images has progressed rapidly in recent years and is now increasingly being applied to and benchmarked with datasets containing remotely sensed imagery. A common benchmark case is to evaluate SSL pre-trained model embeddings on datasets of remotely sensed imagery with small patch sizes, e.g., 32 × 32 pixels, whereas standard SSL pre-training takes place with larger patch sizes, e.g., 224 × 224. Furthermore, pre-training methods tend to use different image normalization preprocessing steps depending on the dataset. In this paper, we show, across seven satellite and aerial imagery datasets of varying resolution, that by simply following the preprocessing steps used in pre-training (precisely, image sizing and normalization methods), one can achieve significant performance improvements when evaluating the extracted features on downstream tasks -- an important detail overlooked in previous work in this space. We show that by following these steps, ImageNet pre-training remains a competitive baseline for satellite imagery based transfer learning tasks -- for example we find that these steps give +32.28 to overall accuracy on the So2Sat random split dataset and +11.16 on the EuroSAT dataset. Finally, we report comprehensive benchmark results with a variety of simple baseline methods for each of the seven datasets, forming an initial benchmark suite for remote sensing imagery.\footnote{Experimental code, datasets, and model checkpoints will be made available in the TorchGeo library at \url{https://github.com/microsoft/torchgeo} and are currently hosted at \url{https://github.com/isaaccorley/resize-is-all-you-need}}
\end{abstract}

\section{Introduction}
\label{sec:intro}
With increasing frequency, self-supervised learning (SSL) models, foundation models, and transfer learning methods have been applied to remotely sensed imagery~~\cite{lacoste2021toward, mai2023opportunities, fuller2022transfer, nguyen2023climax, cha2023billion, tseng2023lightweight, fuller2022satvit, corley2022supervising, manas2021seasonal, reed2022scale, cong2022satmae, wang2022empirical, wang2022ssl4eo, heidler2023self, sun2022ringmo, mikriukov2022deep, neumann2020training}. As such, rigorous benchmarks are needed to identify the strengths and weaknesses in the proposed methods.

A commonly used benchmark in any transfer learning setup is the use of embeddings from a model that is pretrained on the ImageNet (ILSVRC2012) dataset~\cite{deng2009imagenet} -- due to both the ease of implementation~\cite{chollet2015keras,torchvision2016} and strong performance when generalizing to unseen data~\cite{huh2016makes}. However, even with fully convolutional neural networks, the size of image inputs to the model is an important factor that should be controlled for at test/inference time. Common large-scale benchmarks libraries like PyTorch Image Models (timm)~\cite{ross_wightman_2023_7618837} and OpenCLIP~\cite{ilharco_gabriel_2021_5143773} provide benchmark results trained at varying image sizes and evaluate at the same sizes as opposed to the original dataset size. Plainly put, models that are pretrained on ImageNet images that have been resized and cropped to a fixed image size (traditionally 224 × 224 or 256 × 256), will produce the most relevant embeddings for transfer learning when they are given the same image size at test time.

\begin{figure}[t!]
  \centering
  \centerline{\includegraphics[width=0.8\textwidth]{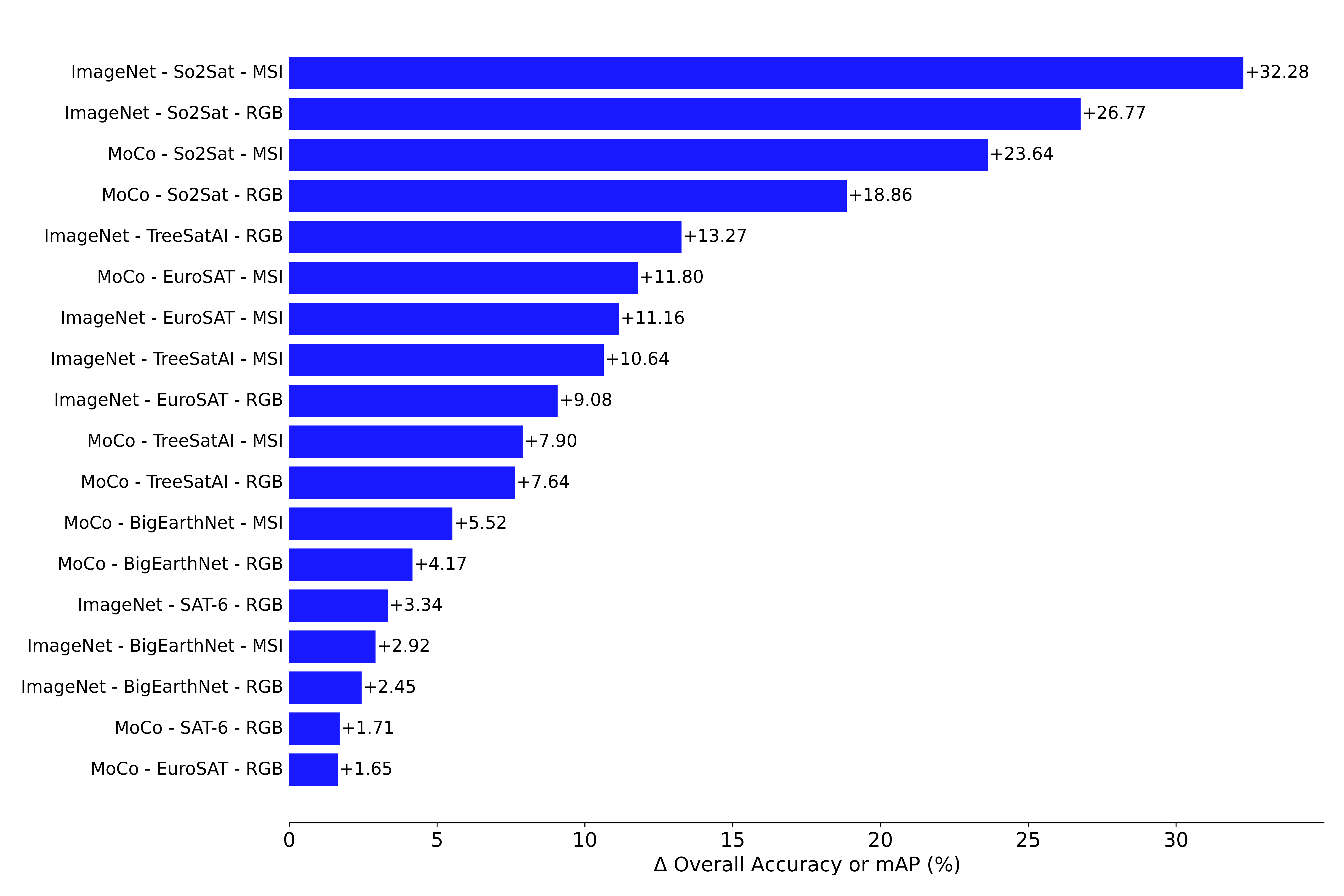}}
\caption{Difference in downstream task metrics, Overall Accuracy (OA) (multiclass) or mean Average Precision (mAP) (multilabel), after resizing images to 224 × 224 from the original, smaller, image size. ImageNet pre-trained models, for example, often are trained with 224 x 224 inputs and therefore do not produce useful embeddings with smaller image patches.}
\label{fig:delta-plot}
\end{figure}

Satellite missions such as Sentinel-2~\cite{drusch2012sentinel} and Landsat-8~\cite{roy2014landsat} capture imagery over the Earth's surface at relatively low spatial resolutions, e.g. 10-60 meters/pixel, compared to the resolution of objects in natural imagery. Because of this, it is common for labeled datasets of remotely sensed imagery to contain images of smaller sizes, e.g. 32 × 32~\cite{zhu2020so2sat}, than traditional image classification datasets. Thus, if images from these datasets are used as-is with ImageNet pretrained models, then the results will be sub-optimal.

A similar story can be told with image normalization methods. A standard preprocessing method for ImageNet pre-trained models is to normalize all values in an image to a $[0,1]$ range then perform channel-wise standardization with ImageNet statistics. However, as remotely sensed imagery usually has a higher bit-depth (or color-depth) than images in standard vision datasets (12 or 16-bit depth vs. 8-bit depth), different image normalizations methods are usually applied. For example, a common method used with Sentinel-2 imagery is to divide all values by 10,000 (to convert the raw sensor values to reflectance values) then use these as inputs in a network~\cite{manas2021seasonal,wang2022ssl4eo}. If images that are normalized with one method are used with a network that is pre-trained under a different normalization method, then the results will also be sub-optimal.

We demonstrate that it is vital to consider how an embedding model was trained when using it for transfer learning on downstream remote sensing tasks. For example, through simple bilinear upsampling of input images from 64 × 64 to 224 × 224 on the EuroSAT RGB dataset~\cite{helber2019eurosat}, we find that accuracy of the embeddings generated by a ImageNet pretrained ResNet-50~\cite{he2016deep} increases from 0.82 to 0.91. Similarly, performing a channel-wise standardization instead of re-scaling the image values to represent reflectance results in a performance increase from 0.66 to 0.91 (when combined with resizing to 224 × 224). \textbf{Performing these steps correctly gives simple baselines, like ImageNet pre-training, results that are competitive with previously published methods.} Additionally, we benchmark several simple methods, including MOSAIKS~\cite{rolf2021generalizable} and a simple image statistic based feature extraction method, and find that they beat ImageNet and/or remote sensing SSL pretraining methods on several datasets.

While not particularly surprising, our results form a set of strong baselines that can be used to benchmark future methods for self-supervised learning with remotely sensed imagery against. Further, our experimental setup is open-sourced and can be easily appended to as the community focuses on different geospatial machine learning tasks.

\begin{figure*}[ht]
    \center
    \includegraphics[width=0.9\textwidth]{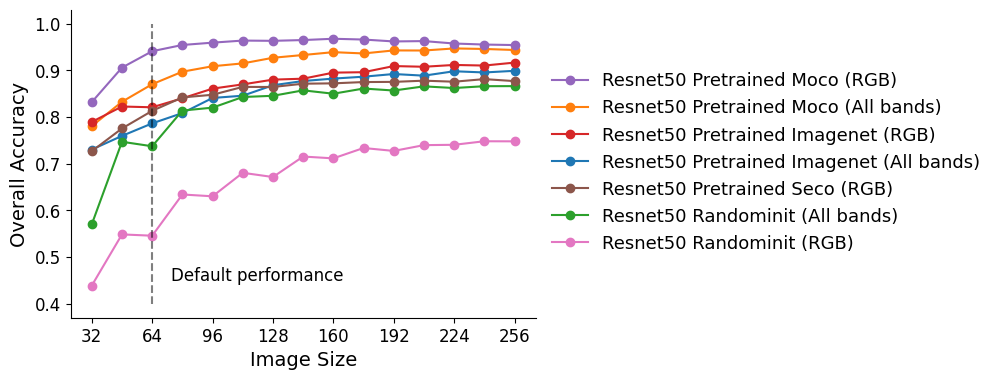}
    \caption{The effect of input image size on EuroSAT downstream performance (overall accuracy) across different ResNet models. By default, EuroSAT images are 64 × 64 pixels, however resizing to larger image sizes before embedding increases downstream accuracy under a KNN ($k=5$) classification model in all cases.}
    \label{fig:eurosat_size_vs_performance}
\end{figure*}

Our main contributions are as follows:
\begin{itemize}
\item We propose a set of strong baseline methods for remote sensing scene classification -- including an ImageNet pretrained ResNet-50, random convolutional features (RCF), and a simple image statistic feature extraction method -- that outperform self-supervised pretrained models on several datasets. We have implemented these methods into the open source TorchGeo library~\cite{stewart2022torchgeo} (see Appendix \ref{sec:code}).

\item We present a set of benchmark results across seven geospatial machine learning datasets commonly used as downstream tasks for testing pre-trained model performance with our baseline methods.

\item We demonstrate the importance of proper resizing and normalization of images for optimal performance and fair comparisons in geospatial machine learning benchmarks.  
\end{itemize}

\subsection{Related Work}
\label{sec:related-work}
Recent works have shown that while many new deep learning architectures claim to achieve state-of-the-art performance due to their proposed novel model design, they in fact only do so because of inconsistencies in training strategies and hyperparameters when comparing to baselines and prior methods. Bello et al.~\cite{bello2021revisiting} explored that by simply retraining with recent training techniques and tricks, the original ResNet~\cite{he2016deep} architecture significantly outperforms its own previous baselines and reaches a competitive top-1 ImageNet accuracy. Du et al.~\cite{du2021revisiting} concluded the same findings for 3D ResNets~\cite{tran2015learning} for video recognition tasks. Goyal et al.~\cite{goyal2021revisiting} examined the similar effects for numerous architectures in the 3D point cloud classification field. Finally, Musgrave et al.~\cite{musgrave2020metric} repeat the same idea for metric learning methods. In other words, when all models are on the same playing field, performance gains from past methods over strong baselines tend to become insignificant. 

Previous papers that explore the effect of resizing inputs on performance in convolutional neural networks include Richter et al.~\cite{richter2021input} and Touvron et al.~\cite{touvron2019fixing}. Both papers investigate different experimental setups by varying training and testing at different image sizes and empirically show that increasing the image size during inference improves performance which begins to saturate around an image size of 256 × 256. However, both works strictly explore natural images only with ImageNet pretraining as opposed to remotely sensed imagery, as is the objective of this paper. Wang et al.~\cite{wang2022ssl4eo} provide the closest evidence of this case for remote sensing data by performing a short experiment reporting linear probing results showing a boost in performance while increasing the input image size.

\begin{table}[t]
    \parbox{.48\linewidth}{
    \captionof{table}{Results on the EuroSAT dataset~\cite{helber2019eurosat} for multiclass classification using KNN ($k=5$). We report Overall Accuracy (OA) for both RGB and all MSI bands. We compare to fine-tuned performance of several SSL methods taken from their respective papers. *The Scale-MAE result uses a KNN-5 and is comparable to the other KNN results.}
    \resizebox{0.49\textwidth}{!}{%
    \begin{tabular}{@{}ccccc@{}}
    \toprule
    \textbf{Model} &
    \textbf{Weights} &
    \textbf{Size} &
    \textbf{RGB} &
    \textbf{MSI} \\
    \toprule
    
    \multirow{2}{*}{ResNet50} & \multirow{2}{*}{MoCo} & 64 & \textit{94.11} & 81.85 \\
     &  & 224 & \textbf{95.76} & \textbf{93.65} \\ \midrule
    \multirow{2}{*}{ResNet50} & \multirow{2}{*}{ImageNet} & 64 & 82.09 & 78.65 \\
     &  & 224 & 91.17 & 89.81 \\ \midrule
    \multirow{2}{*}{ResNet50} & \multirow{2}{*}{Random} & 64 & 59.92 $\pm$ 0.34 & 75.10 $\pm$ 0.23 \\
     &  & 224 & 73.76 $\pm$ 0.53 & 87.19 $\pm$ 0.81 \\ \midrule
    \multirow{2}{*}{RCF} & \multirow{2}{*}{Random} & 64 & 78.85 $\pm$ 0.33 & 87.56 $\pm$ 0.35 \\
     &  & 224 & 76.90 $\pm$ 0.33 & 87.41 $\pm$ 0.12 \\ \midrule
     \multirow{2}{*}{RCF} & \multirow{2}{*}{Empirical} & 64 & 81.47 $\pm$ 0.08 & \textit{91.10 $\pm$ 0.11} \\
     &  & 224 & 77.88 $\pm$ 0.08 & 90.14 $\pm$ 0.15 \\ \midrule
    Image Stat. & - & 64 & 76.94 & 89.56 \\ \midrule \midrule
    ViT-L & Scale-MAE~\cite{reed2022scale} & 64 & 96.00* & - \\
    ResNet18 & GASSL~\cite{ayush2021geography} & 64 & 89.51 &  - \\
    ResNet18 & SeCo~\cite{manas2021seasonal} & 64 & 93.14 &  - \\
    ViT-L & SatMAE~\cite{cong2022satmae} & 224 & 98.94 &  - \\
    \bottomrule
    \end{tabular}%
    }
    \label{tab:results-eurosat}
    }
    \qquad
    \parbox{.48\linewidth}{
    \captionof{table}{
    Results on the SAT-6 dataset~\cite{basu2015deepsat} for multiclass classification using KNN ($k=5$). We report Overall Accuracy (OA) and compare to the fully-supervised performance of DeepSAT and DeepSATv2 models taken from their respective papers.}
    \resizebox{0.49\textwidth}{!}{
    \begin{tabular}{ccccc}
        \toprule
        \multicolumn{1}{c}{\textbf{Model}} &
        \multicolumn{1}{c}{\textbf{Weights}} &
        \multicolumn{1}{c}{\textbf{Size}} &
        \multicolumn{1}{c}{\textbf{OA}} \\
        \toprule
        \multirow{2}{*}{ResNet50} & \multirow{2}{*}{MoCo} & 34 & 98.15 \\
         &  & 224 & \textit{99.86} \\
        \midrule
        \multirow{2}{*}{ResNet50} & \multirow{2}{*}{ImageNet} & 34 & 96.55 \\
         &  & 224 & \textbf{99.89} \\
        \midrule
        \multirow{2}{*}{ResNet50} & \multirow{2}{*}{Random} & 34 & 91.64 $\pm$ 0.66 \\
         &  & 224 & 98.57 $\pm$ 0.08 \\
        \midrule
        \multirow{2}{*}{RCF} & \multirow{2}{*}{Random} & 34 & 99.40 $\pm$ 0.06 \\
         &  & 224 & 99.29 $\pm$ 0.07 \\
        \midrule
        \multirow{2}{*}{RCF} & \multirow{2}{*}{Empirical} & 34 & 99.65 $\pm$ 0.02 \\
         &  & 224 & 98.85 $\pm$ 0.06 \\
        \midrule
        Image Stat. & - & 28 & 99.60 \\
        \midrule \midrule
        DeepSat~\cite{basu2015deepsat} & Sup. & 28 & 93.92 \\
        DeepSatv2~\cite{liu2020deepsat} & Sup. & 28 & 99.84 \\
        \bottomrule
    \end{tabular}
    }
    \label{tab:results-sat6}
    }
\end{table}

\section{Methods}
\label{sec:methods}
In this study we extract feature representations (or embeddings) from remotely sensed image datasets using a variety of methods (described below) while varying the image preprocessing steps. Specifically, we vary the image size that is passed through to the feature extractor using Pytorch's~\cite{Paszke_PyTorch_An_Imperative_2019} \verb|torch.nn.functional.interpolate| implementation with bilinear interpolation, and we vary the image normalization method between channel-wise standardization (i.e. the default practice for most ImageNet pretrained models), converting the input image values into a reflectance value (i.e. the default practice for most Sentinel-2 pretrained models), min-max normalization, or method specific normalizations (e.g. the percentile normalization from~\citep{manas2021seasonal}). In datasets that have multispectral information we run experiments using only the RGB channels, as well as all the channels (MSI)\footnote{Note that for processing multispectral (MSI) imagery through ImageNet pretrained ResNets, we repeat the RGB weights in the first convolutional layer to account for the additional input bands. For SSL4EO MSI pretrained ResNets, we zero-pad channels to account for any bands not made available in datasets.}.

We extract feature representations using the following methods:
\begin{description}
    \item[ResNet-50 Random init.~\cite{he2016deep}] A vanilla ResNet-50 with random weight initialization  (following the default torchvision settings). The features generated by this and the following two ResNet-50 models are produced by the final global average pool operation and are 2048-dimensional.

    \item[ResNet-50 ImageNet~\cite{deng2009imagenet}]  A ResNet-50 that is pretrained on ImageNet with images of size 224x224 (default torchvision pretrained weights).

    \item[ResNet-50 SSL4EO~\cite{wang2022ssl4eo}] A ResNet-50 that is pretrained using the MoCo-v2~\cite{he2020momentum,chen2020improved} self-supervised learning method on the SSL4EO dataset with 224x224 images.

    \item[RCF (Random)~\cite{rolf2021generalizable}] A feature extraction method that consists of projecting the input to a lower dimensional space using random convolutional features (RCF). We use the implementation from TorchGeo with 512 convolutional filters and a 3x3 kernel size. In the results we refer to this method as RCF with random weights.

    \item[MOSAIKS / RCF (Empirical)~\cite{rolf2021generalizable}] A feature extraction method similar to RCF but that initializes the weights using ZCA whitened patches sampled randomly from the training set. We use the implementation from TorchGeo with 512 convolutional filters and a 3x3 kernel size. In the results we refer to this method as RCF with empirical weights.

    \item[Image Statistics] A hand crafted baseline method that consists of simply computing per-channel pixel statistics from the imagery. Given an image we compute the mean, standard deviation, minimum, and maximum value for each band and concatenate these into a simple $4c$-dimensional feature representation, where $c$ is the number of input channels.
\end{description}

\subsection{Evaluation}
For evaluating the representation performance of a pretrained model it is common to perform ``linear probing'' on a given downstream task by training a linear model on the representations generated by the pre-trained model and measuring the performance of this linear model. However, this method is implemented very differently between papers -- some papers use data augmentation~\cite{wang2022ssl4eo} while others don't, and others use a variety of different optimizers (SGD, Adam, LARS), regularization methods\footnote{For example, by default the Adam optimizer in PyTorch will not apply L2 regularization on the weights of the model (weight decay), while scikit-learn linear models are trained with L2 regularization by default.}, and learning rates / learning rate schedules. Therefore, for fair evaluation we fit a K-Nearest-Neighbors (KNN) model~\cite{cover1967nearest} to extracted features from various datasets, setting $k=5$, as performed similarly in~\cite{reed2022scale,tseng2023lightweight}.

\section{Datasets} \label{sec:data}
The datasets used throughout our experiments were selected particularly due to their original image sizes being small to show the effects of resizing. These datasets are commonly benchmarked without resizing which makes them perfect candidates for quantifying the effects of size vs performance. We also select datasets which are from both low-resolution satellite sources as well as high resolution aerial imagery.

\begin{description}
    \item[EuroSAT] The EuroSAT dataset~\cite{helber2019eurosat} is a land cover classification dataset of patches extracted from multispectral Sentinel-2~\cite{drusch2012sentinel} imagery. The dataset contains 27,000 64 × 64 10m spatial resolution images with 13 bands and labels for 10 land cover categories. We use the dataset splits defined in Neumann et al.~\cite{neumann2019domain}.

    \item[SAT-6] The SAT-6 dataset~\cite{basu2015deepsat} is a land cover classification dataset of patches extracted from aerial imagery from the National Agriculture Imagery Program (NAIP)~\cite{naip}. The dataset contains 405,000 28 × 28 RGBN patches at 1m spatial resolution and labels for 6 land cover categories. We use the train and test splits provided with the dataset.

    \item[So2Sat] The So2Sat dataset~\cite{zhu2020so2sat} is a local climate zone (LCZ) classification dataset of patches extracted from Sentinel-1 and Sentinel-2 imagery. For our experiments we only utilize the Sentinel-2 bands. The dataset contains 400,673 multispectral patches with 10 bands and at 10m spatial resolution. Each patch is of size 32 × 32 and contains a single label from 17 total LCZ categories. We use the train and test splits from the Random and Culture-10 sets provided with the dataset. 

    \item[BigEarthNet] The BigEarthNet dataset~\cite{sumbul2019bigearthnet} is a multi-label land cover classification dataset of patches extracted from multispectral Sentinel-2 imagery. The dataset contains 590,326 120 × 120 10m spatial resolution images with 12 bands and labels for 19 land cover categories. We use the splits provided with the dataset and defined in~\cite{sumbul2021bigearthnet}.

    \item[TreeSatAI] The TreeSatAI dataset~\cite{ahlswede2023treesatai} is a multi-sensor, multilabel tree species classification dataset of patches extracted from aerial and multispectral Sentinel-1~\cite{torres2012gmes} and Sentinel-2 imagery.  For our experiments we only utilize the Sentinel-2 bands. The dataset contains 50,381 10m spatial resolution images with 12 spectral bands, which are available in 6 × 6 or 20 × 20 sizes, and labels for 20 tree species categories. We use the train and test splits provided with the dataset.

    \item[UC Merced] The UC Merced (UCM) dataset~\cite{yang2010bag} is a land use classification dataset that consists of 2,100 256 × 256 pixel aerial RGB images over 21 target classes. We use the train/val/test splits defined in Neumann et al.~\cite{neumann2019domain}.

    \item[RESISC45] The RESISC45 dataset~\cite{cheng2017remote} is a scene classification dataset that consists of 45 scene classes and 31,500 256 × 256 pixel aerial RGB images extracted from Google Earth. We use the dataset splits defined in Neumann et al.~\cite{neumann2019domain}.
\end{description}

\section{Results and Discussion}
\label{sec:results}
\begin{table}[t]
\centering
\caption{Results on the So2Sat dataset~\cite{zhu2020so2sat} for multiclass classification using KNN ($k=5$). We report Overall Accuracy (OA) for both RGB and all MSI bands and for both the \textit{Random} and \textit{Culture-10} splits. We compare to both fully-supervised and linear probing results for several SSL methods.}
\resizebox{0.9\textwidth}{!}{%
\begin{tabular}{ccccccc}
\toprule
\multicolumn{3}{c}{\textbf{}} &
\multicolumn{2}{c}{\textbf{Random}} &
\multicolumn{2}{c}{\textbf{Culture-10}} \\
\cmidrule(lr){4-5}\cmidrule(lr){6-7}

\textbf{Model} &
\textbf{Weights} &
\textbf{Size} &
\textbf{RGB} &
\textbf{MSI} &
\textbf{RGB} &
\textbf{MSI} \\
\toprule

\multirow{2}{*}{ResNet50} & \multirow{2}{*}{MoCo} & 34 & 75.07 & 72.51 & 51.45 & 49.36 \\
 &  & 224 & \textbf{93.93} & \textbf{96.15} & \textbf{56.03} & \textbf{53.54} \\
\midrule
\multirow{2}{*}{ResNet50} & \multirow{2}{*}{ImageNet} & 34 & 66.21 & 56.18 & 47.76 & 42.11 \\
 &  & 224 & \textit{92.99} & 88.46 & \textit{54.53} & \textit{50.32} \\
\midrule
\multirow{2}{*}{ResNet50} & \multirow{2}{*}{Random} & 34 & 46.19 $\pm$ 0.19 & 55.06 $\pm$ 0.35 & 29.10 $\pm$ 0.30 & 35.47 $\pm$ 0.18 \\
 &  & 224 & 71.74 $\pm$ 1.87 & 84.10 $\pm$ 0.32 & 34.16 $\pm$ 0.23 & 45.68 $\pm$ 0.50 \\
\midrule
\multirow{2}{*}{RCF} & \multirow{2}{*}{Random} & 34 & 72.67 $\pm$ 0.45 & 89.40 $\pm$ 0.14 & 30.92 $\pm$ 0.11 & 45.23 $\pm$ 0.33 \\
 &  & 224 & 74.22 $\pm$ 0.44 & 89.72 $\pm$ 0.11 & 31.19 $\pm$ 0.21 & 45.36 $\pm$ 0.36 \\
\midrule
\multirow{2}{*}{RCF} & \multirow{2}{*}{Empirical} & 34 & 71.00 $\pm$ 0.32 & \textit{95.37 $\pm$ 0.06} & 35.32 $\pm$ 0.45 & 47.63 $\pm$ 0.10 \\
 &  & 224 & 51.66 $\pm$ 0.46 & 95.20 $\pm$ 0.02 & 27.36 $\pm$ 0.24 & 44.98 $\pm$ 0.16 \\
\midrule
Image Stat. & - & 32 & 83.84 & 91.09 & 38.36 & 47.93 \\
\midrule\midrule
ResNet50 & MoCo~\cite{wang2022ssl4eo} & 224 & - & - & - & 61.80 \\
ResNet50 & DINO~\cite{caron2021emerging} & 224 & - & - & - & 57.00 \\
ViT-S & DINO~\cite{caron2021emerging} & 224 & - & - & - & 62.50 \\
ViT-S & MAE~\cite{he2022masked} & 224 & - & - & - & 60.00 \\
ResNet50 & Sup.~\cite{wang2022ssl4eo} & 224 & - & - & - & 57.50 \\
ViT-S & Sup.~\cite{wang2022ssl4eo} & 224 & - & - & - & 59.30 \\
\bottomrule
\end{tabular}%
}
\label{tab:results-so2sat-full}
\end{table}

\subsection{Fair Comparisons to ImageNet Pretraining}
As stated in Section~\ref{sec:related-work}, prior research has shown the significance of resizing images during testing for ImageNet pretrained models. To emphasize this, we perform a short experiment comparing features extracted from the EuroSAT~\cite{helber2019eurosat} dataset using a ResNet-18 pretrained with both the Seasonal Contrast (SeCo) method~\cite{manas2021seasonal} and ImageNet. For fair evaluation, we compute downstream task results at the original image size 64 × 64 and resized to 224 × 224 with KNN and linear probe methods. For linear probing we utilize the exact same experimental setup and script as in \cite{manas2021seasonal} while only adding a resize transformation. As seen in Table~\ref{tab:results-eurosat-seco-vs-imagenet}, depending on the model used for evaluation, one pretraining method can appear better than another. Furthermore, while increasing the image size improves performance for both methods, it does not improve equally. When reading the linear probing results in ~\cite{manas2021seasonal}, one would assume that the SSL pretrained model clearly outperforms ImageNet pretraining. However, as we can see, this is not the case, and further investigation are needed. Further, in Table \ref{tab:results-ucm}, we observe that an ImageNet pretrained model outperforms the best reported results in SatMAE~\cite{cong2022satmae} in the same experimental setup.

\begin{figure}[ht!]

\centering
\includegraphics[width=.33\textwidth]{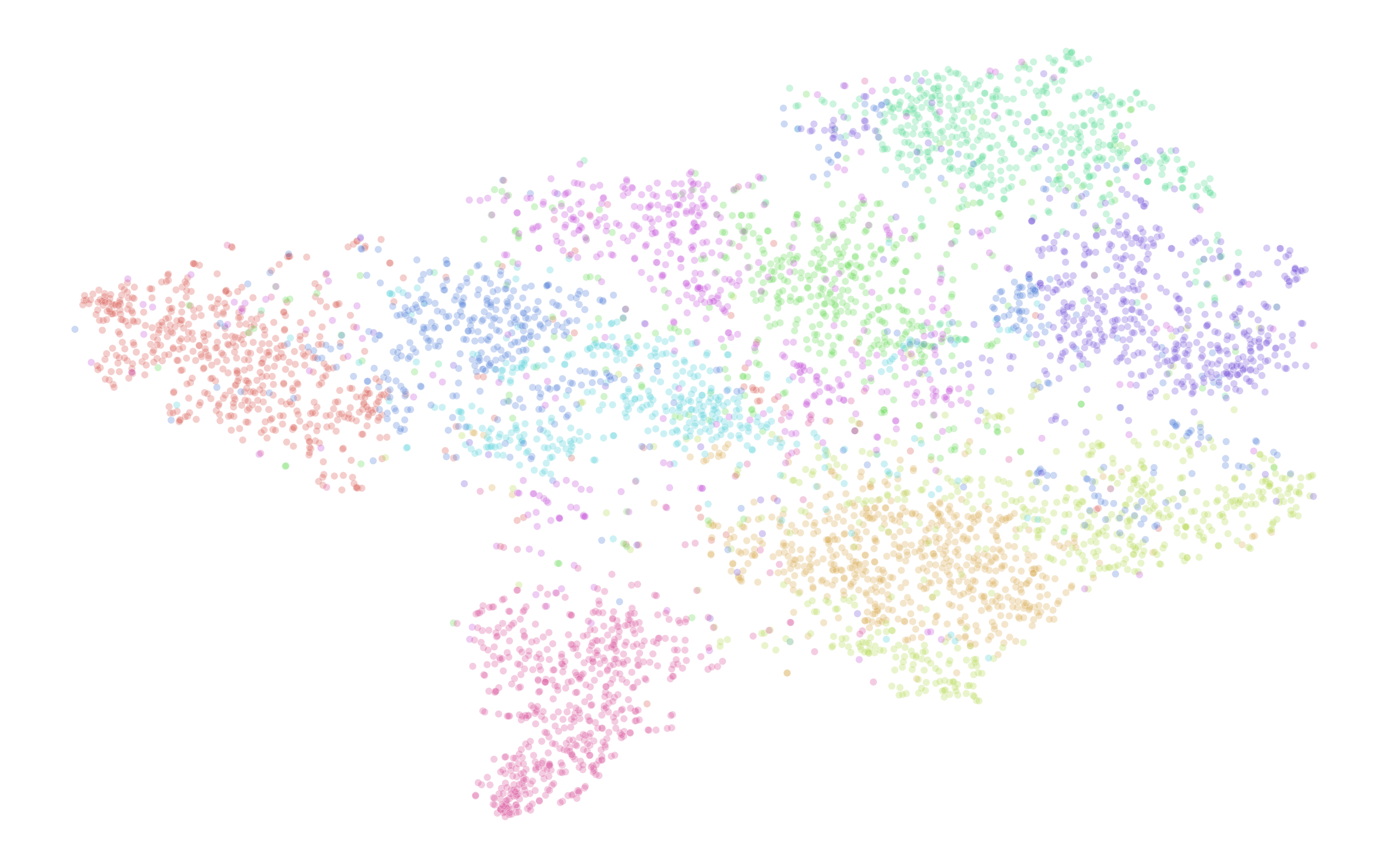}\hfill
\includegraphics[width=.33\textwidth]{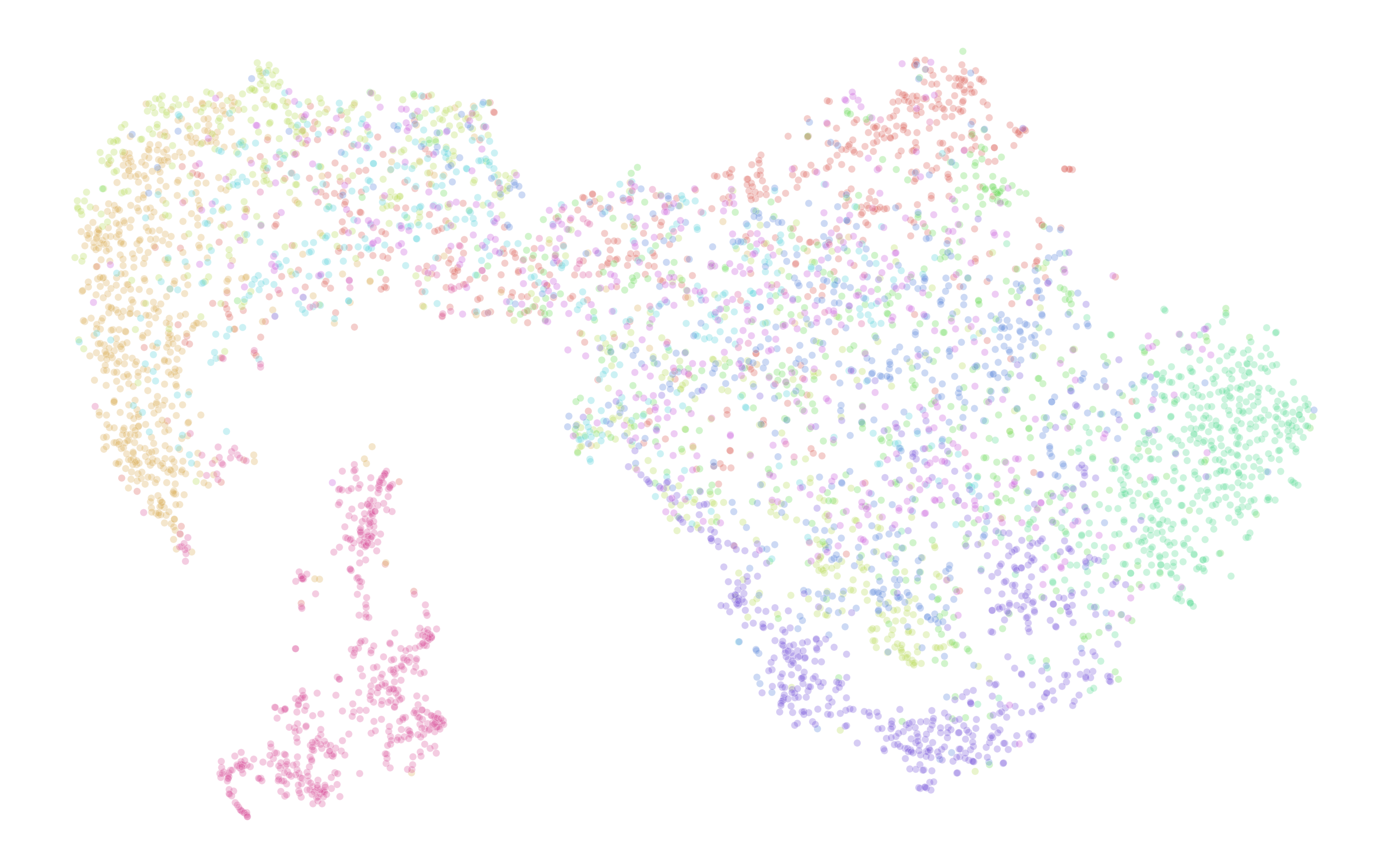}\hfill
\includegraphics[width=.33\textwidth]{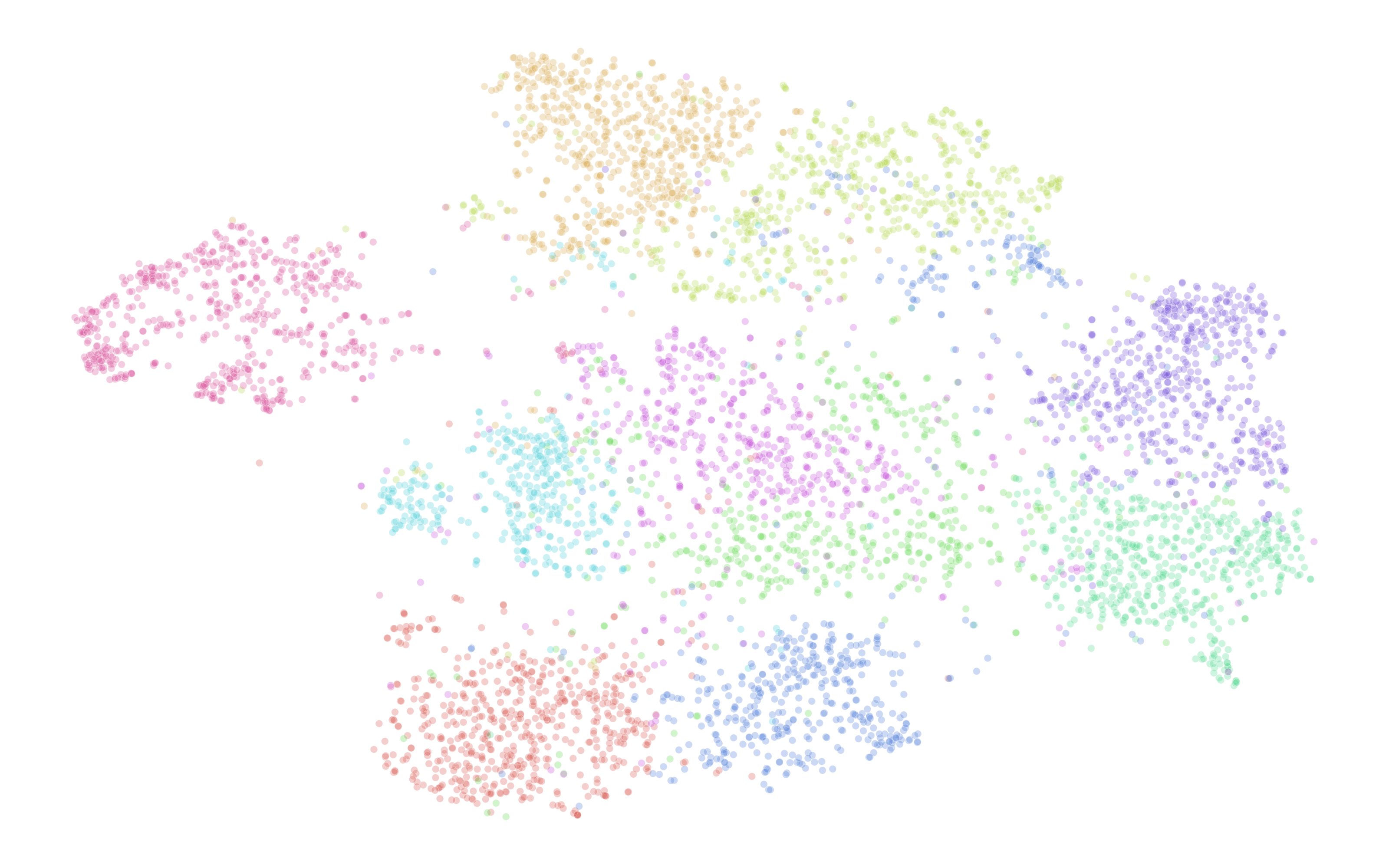}
\caption{t-SNE~\cite{van2008visualizing} plots of EuroSAT test set embeddings extracted using a ResNet50 pretrained on ImageNet with different preprocessing. (left to right: 32 × 32 with normalization, 224 × 224 without normalization, 224 × 224 with normalization)}
\label{fig:tsne}
\end{figure}

\subsection{Image Size vs. Performance}
Figure \ref{fig:eurosat_size_vs_performance} shows how the performance of a variety ResNet-50 models varies with input image size on the EuroSAT dataset when using just the RGB bands vs. all spectral bands as input. We observe in all cases that the default dataset image size (64 × 64 pixels) does not result in optimal performance. For example, resizing from 64 × 64 to 256 × 256 results in a 10 point increase in accuracy in a ResNet-50 that is pretrained on ImageNet. In Tables \ref{tab:results-eurosat}-\ref{tab:results-treesatai} we report performance from each method at the native resolution of the dataset and after resizing each image to 224x224 and observe performance improvements across all methods in nearly all cases.

To visualize the effects of resizing (and standard normalization), in Figure~\ref{fig:tsne} we show t-SNE~\cite{van2008visualizing} plots of EuroSAT RGB features extracted using a ResNet-50 pretrained on ImageNet. The the plot shows that EuroSAT classes are clearly separable at an input size of 224 x 224 while only partially separable at 32 x 32. Additionally, when resizing but not using any normalization, there are no clear clusters corresponding to the dataset classes. While we use a NVIDIA DGX server with 2x A100 GPUs to increase the speed of our benchmarks, we note that none of these methods actually require a GPU to perform inference or KNN classification on extracted features.

\begin{table}[ht!]
\centering
\caption{Results on the BigEarthNet dataset~\cite{sumbul2019bigearthnet} for 19-class multilabel classification using KNN ($k=5$). We report overall F1 score, and overall mean average precision (mAP). For reference, we compare to the fully supervised S-CNN as well as fine-tuned results from the GASSL, SeCo, and SatMAE SSL methods.}
\resizebox{0.9\textwidth}{!}{%
\begin{tabular}{ccccccc}
\toprule
\multicolumn{3}{c}{\textbf{}} &
\multicolumn{2}{c}{\textbf{RGB}} &
\multicolumn{2}{c}{\textbf{MSI}} \\
\cmidrule(lr){4-5}\cmidrule(lr){6-7}

\textbf{Model} &
\textbf{Weights} &
\textbf{Size} &
\textbf{F1} &
\textbf{mAP} &
\textbf{F1} &
\textbf{mAP} \\
\toprule

\multirow{2}{*}{ResNet50} & \multirow{2}{*}{MoCo} & 120 & \textit{68.99} & \textit{70.65} & 63.61 & 64.64 \\
 &  & 224 & \textbf{72.56} & \textbf{74.81} & 68.33 & 70.17 \\
\midrule
\multirow{2}{*}{ResNet50} & \multirow{2}{*}{ImageNet} & 120 & 65.38 & 66.62 & 62.61 & 62.96 \\
 &  & 224 & 67.47 & 69.07 & 65.04 & 65.88 \\
\midrule
\multirow{2}{*}{ResNet50} & \multirow{2}{*}{Random} & 120 & 52.34 $\pm$ 0.22 & 52.63 $\pm$ 0.19 & 60.48 $\pm$ 0.34 & 61.17 $\pm$ 0.50 \\
 &  & 224 & 57.05 $\pm$ 1.02 & 57.61 $\pm$ 1.13 & 64.94 $\pm$ 0.25 & 66.31 $\pm$ 0.32 \\
\midrule
\multirow{2}{*}{RCF} & \multirow{2}{*}{Random} & 120 & 54.48 $\pm$ 0.26 & 53.94 $\pm$ 0.26 & 69.98 $\pm$ 0.20 & 72.01 $\pm$ 0.28 \\
 &  & 224 & 54.37 $\pm$ 0.28 & 53.74 $\pm$ 0.23 & 70.06 $\pm$ 0.21 & 72.12 $\pm$ 0.29 \\
\midrule
\multirow{2}{*}{RCF} & \multirow{2}{*}{Empirical} & 120 & 57.40 $\pm$ 0.22 & 57.22 $\pm$ 0.23 & \textit{73.31 $\pm$ 0.14} & \textit{76.18 $\pm$ 0.19} \\
 &  & 224 & 53.36 $\pm$ 0.23 & 52.90 $\pm$ 0.22 & \textbf{73.41 $\pm$ 0.13} & \textbf{76.29 $\pm$ 0.15} \\
\midrule
Image Stat. & - & 120 & 61.67 & 62.00 & 69.42 & 71.29 \\
\midrule \midrule
S-CNN & BigEarthNet~\cite{sumbul2019bigearthnet} & 120 & 67.59 & - & 70.98 & - \\
ResNet50 & GASSL~\cite{ayush2021geography} & 120 & - & 80.20 & - & - \\
ResNet50 & SeCo~\cite{manas2021seasonal} & 120 & - & 82.62 & - & - \\
ViT-L & SatMAE~\cite{cong2022satmae} & 224 & - & 82.13 & - & - \\
\bottomrule
\end{tabular}%
}
\label{tab:results-bigearthnet}
\end{table}
\begin{table}[t!]
\centering
\caption{Results on the TreeSatAI dataset~\cite{ahlswede2023treesatai} for multilabel classification using KNN ($k=5$). We report overall F1 score and mean average precision mAP. We compare to the fully-supervised LightGBM performance and fine-tuned Presto SSL method.}
\resizebox{0.9\textwidth}{!}{%
\begin{tabular}{ccccccc}
\toprule
 &  & \multicolumn{1}{l}{} &
\multicolumn{2}{c}{\textbf{RGB}} &
\multicolumn{2}{c}{\textbf{MSI}} \\
\cmidrule(lr){4-5}\cmidrule(lr){6-7}
 
\textbf{Model} &
\textbf{Weights} &
\textbf{Size} &
\textbf{F1} &
\textbf{mAP} &
\textbf{F1} &
\textbf{mAP} \\
\toprule

\multirow{2}{*}{\textbf{ResNet50}} & \multirow{2}{*}{MoCo} & 34 & 29.21 & 29.93 & 37.65 & 36.24 \\
 &  & 224 & 37.68 & \textit{37.57} & 45.18 & 44.14 \\
\midrule
\multirow{2}{*}{ResNet50} & \multirow{2}{*}{ImageNet} & 34 & 27.69 & 27.30 & 32.07 & 30.69 \\
 &  & 224 & \textbf{40.37} & \textbf{40.58} & 42.00 & 41.33 \\
\midrule
\multirow{2}{*}{ResNet50} & \multirow{2}{*}{Random} & 34 & 29.37 $\pm$ 0.42 & 29.08 $\pm$ 0.18 & 36.47 $\pm$ 0.34 & 34.73 $\pm$ 0.15 \\
 &  & 224 & 35.42 $\pm$ 0.33 & 34.75 $\pm$ 0.43 & 49.09 $\pm$ 0.83 & 48.48 $\pm$ 0.89 \\
\midrule
\multirow{2}{*}{RCF} & \multirow{2}{*}{Random} & 34 & 33.15 $\pm$ 0.21 & 32.15 $\pm$ 0.09 & 52.24 $\pm$ 0.35 & 51.83 $\pm$ 0.33 \\
 &  & 224 & 32.37 $\pm$ 0.20 & 31.29 $\pm$ 0.18 & 52.49 $\pm$ 0.17 & 51.99 $\pm$ 0.43 \\
\midrule
\multirow{2}{*}{RCF} & \multirow{2}{*}{Empirical} & 34 & 31.70 $\pm$ 0.06 & 31.13 $\pm$ 0.17 & \textbf{56.00 $\pm$ 0.04} & \textbf{56.08 $\pm$ 0.25} \\
 &  & 224 & 28.93 $\pm$ 0.47 & 28.50 $\pm$ 0.23 & \textit{55.60 $\pm$ 0.13} & \textit{55.77 $\pm$ 0.29} \\
\midrule
Image Stat. & - & 20 & \textit{38.39} & 37.19 & 51.97 & 51.56 \\
\midrule\midrule
LightGBM~\cite{ke2017lightgbm} & - & 20 & - & - & 52.52 & 61.66 \\
ViT & Presto~\cite{tseng2023lightweight} & 9 & - & - & 50.32 &  67.78 \\
\bottomrule
\end{tabular}%
}
\label{tab:results-treesatai}
\end{table}

\subsection{Benchmarks}

We perform thorough benchmarks using the methods described in Section \ref{sec:methods} on each dataset from Section \ref{sec:data}, using the evaluation metric common to that dataset, in Tables \ref{tab:results-eurosat} through \ref{tab:results-ucm}. In each experiment we fit a non-parametric k-nearest neighbor model with $k=5$ to the train set. For deterministic methods we report a single value calculated over the test set for each dataset, while for stochastic methods we report the average $\pm$ the standard deviation of the metric calculated over the test set over 5 runs with different random seeds. We bold the best performing of the baseline methods by column and italicize the second best performing method. Additionally, we show several fine-tuning, linear probing, and fully-supervised baselines from original dataset papers or other SSL remote sensing papers. Note that we perform these comparisons not with the goal of outperforming them but for transparency of the difference in performance in representation ability to the state-of-the-art. Finally, we note that our evaluation method is the same as that of Reed et al.~\cite{reed2022scale} and indicate this with an asterisk where appropriate.

For the EuroSAT experiments we show results from GASSL~\cite{ayush2021geography}, SeCo~\cite{manas2021seasonal}, and SatMAE~\cite{cong2022satmae} self-supervised methods that use fine-tuning on top of the pretrained network (as reported by SatMAE). We note that methods which use a (ViT)~\cite{dosovitskiy2020image} model are unable to accept input images with varying sizes and therefore we only report performance from their original training image size.

For the SAT-6 experiments we compare to the performance of the DeepSat~\cite{basu2015deepsat} model proposed in the original SAT-6 dataset paper as well as the DeepSatv2~\cite{liu2020deepsat} model from a follow-up paper.

For the UC Merced experiments, we compare to the performance of SatMAE~\cite{cong2022satmae}, Scale-MAE~\cite{reed2022scale}, and ConvMAE~\cite{gao2022convmae} as reported in the Scale-MAE paper.

\begin{table}[t!]
    \parbox{.48\linewidth}{
    \captionof{table}{Results on the RESISC45 dataset~\cite{cheng2017remote} for multiclass classification using KNN ($k=5$). We report Overall Accuracy (OA) and compare to performance metrics of various remote sensing SSL methods taken from their respective papers. *The Scale-MAE result uses a KNN-5 and is comparable to the other KNN results.}
    \resizebox{0.49\textwidth}{!}{
    \begin{tabular}{ccccc}
        \toprule
        \textbf{Model} &
        \textbf{Weights} &
        \textbf{Size} &
        \textbf{OA} \\
        \toprule
        ResNet-50 & MoCo & 256 & \textit{73.24} \\
        ResNet-50 & ImageNet & 256 & \textbf{77.48} \\
        ResNet-50 & Random & 256 & 36.30 $\pm$ 0.25 \\
        RCF & Random & 256 & 42.29 $\pm$ 0.12 \\
        RCF & Empirical & 256 & 36.15 $\pm$ 0.36 \\
        Image Stat. & - & 256 & 34.03 \\ \midrule
        ViT-L & Scale-MAE~\cite{reed2022scale} & 256 & 85.0 *\\
        ViT-L & SatMAE~\cite{cong2022satmae} & 256 & 77.1* \\
        ViT-L & ConvMAE~\cite{gao2022convmae} & 256 & 78.8* \\
        \bottomrule
    \end{tabular}
    }
    \label{tab:results-resisc45}
    }
    \qquad
    \parbox{0.48\linewidth}{
    \captionof{table}{Results on the UC Merced dataset~\cite{yang2010bag} for multiclass classification using KNN ($k=5$). We report Overall Accuracy (OA) and compare to the linear probing performance of the Scale-MAE, SatMAE, and ConvMAE methods taken from their respective papers. *The Scale-MAE result uses a KNN-5 and is comparable to the other KNN results.}
    \resizebox{0.49\textwidth}{!}{
    \begin{tabular}{@{}cccc@{}}
        \toprule
        \textbf{Model} &
        \textbf{Weights} &
        \textbf{Size} &
        \textbf{OA} \\
        \toprule
        ResNet50 & MoCo & 256 & \textit{85.50} \\
        ResNet50 & ImageNet & 256 & \textbf{90.70} \\
        ResNet50 & Random & 256 & 47.94 $\pm$ 1.07 \\
        RCF & Random & 256 & 52.14 $\pm$ 0.24 \\
        RCF & Empirical & 256 & 56.90 $\pm$ 0.63 \\
        Image Stat. & - & 256 & 47.90 \\ \midrule
        ViT-L & Scale-MAE~\cite{reed2022scale} & 256 & 85.1* \\
        ViT-L & SatMAE~\cite{cong2022satmae} & 256 & 84.2* \\
        ViT-L & ConvMAE~\cite{gao2022convmae} & 256 & 81.7* \\
        \bottomrule
    \end{tabular}
    }
    \label{tab:results-ucm}
    }
\end{table}

Our results show the following:
\begin{itemize}
    \item SSL4EO MoCo-v2 pretrained weights have the best overall performance across downstream tasks. They rank in the top-2 methods by performance for 6 out of the 7 RGB datasets, and 3 out of 5 multispectral datasets.
    \item The Scale-MAE pretrained model performs the best in the EuroSAT and RESISC45 datasets, however is outperformed by ImageNet pretraining in the UCM dataset.
    \item MOSAIKS (i.e. RCF with empirical weights) is a very strong baseline on the multispectral datasets and ranks in the top 2 methods by performance for 4 out of the 5 multispectral datasets (counting the Random and Culture-10 splits of So2Sat as seperate datasets).  
    \item The image statistic baseline outperforms ImageNet pretrained models on all but one of the multispectral datasets (and it is 0.25\% lower than ImageNet in this case).
    \item In SAT-6 experiments, all methods except for the randomly initialized ResNet-50 achieve greater than 99\% accuracy. Even the image statistic baseline achieves a 99.6\% overall accuracy. This suggests that the dataset is too simple to be used as a benchmark for comparing models as it will be difficult to observe statistically significant changes in accuracy between 99.6\% (any result worse than this would suggest a model that is less expressive than simply extracting image statistics) and 100\%. Nevertheless, future work could explore this dataset in other settings, such as few-shot learning.
    \item Resizing images does not result in significantly changed downstream performance with the RCF methods (as compared to the ResNet based models). We hypothesize that this method is largely scale invariant -- however leave further experiments (such as varying convolutional size with input size, etc.) to future work.
    \item Out of the five datasets with multispectral information, adding the additional multispectral bands to the RGB bands degrades ResNet-50 ImageNet pretrained performance in two cases. However, in all cases, adding multispectral information increases the ResNet-50 random initialized performance. This further highlights the difference in distributions between ImageNet, natural imagery, and remotely sensed imagery.
    \item In the So2Sat dataset, switching from the Random set to the Culture-10 set decreases the accuracy of RCF methods more than the pre-trained models. We hypothesize that this is because the Culture-10 set tests geographic generalization, and RCF will only be able to use color/texture from the train set while the pre-trained models could potentially group similar patches across sets to similar feature representations.  
\end{itemize}

\begin{table}[ht]
\centering
\caption{Comparison of SeCo~\cite{manas2021seasonal} vs. ImageNet pretraining on the EuroSAT validation set. We show Overall Accuracy results for both KNN and linear probe at different image sizes.}
\resizebox{0.6\textwidth}{!}{%
\begin{tabular}{ccccc}
\toprule
\textbf{Size} &
\textbf{Weights} &
\textbf{KNN ($\bm{k=3}$)} &
\textbf{KNN ($\bm{k=10}$)} &
\textbf{Linear Probe} \\
\toprule

\multirow{2}{*}{64} & SeCo & 84.04 & 84.11 & \textbf{93.14} \\
 & ImageNet & \textbf{85.39} & \textbf{85.20} & 86.44 \\
\midrule
\multirow{2}{*}{224} & SeCo & 86.57 & 85.63 & \textbf{96.30} \\
 & ImageNet & \textbf{90.54} & \textbf{90.63} & 93.13 \\

\bottomrule
\end{tabular}%
}
\label{tab:results-eurosat-seco-vs-imagenet}
\end{table}

\section{Best Practices} \label{sec:discussion}
To recap, below is a list of best practices we believe all remote sensing pre-training research should include in their analyses. While these may seem obvious, it is critical to follow these guidelines to produce accurate and transparent benchmarks for understanding the strengths and weaknesses of proposed methods to the community.

\begin{enumerate}
    \item \textbf{Always compare to simple baseline}: Performance across datasets can be misleading, therefore always compare a simple and effective baseline. We recommend an ImageNet pretrained model, random convolutional features, and image statistics.
    \item \textbf{Resize \& Normalize}: Resize and normalize inputs to the same parameters as during training, for all methods being compared. For example, when comparing to ImageNet pretrained models perform min/max normalization on inputs to the range $[0, 1]$, perform channel-wise standardization to scale inputs to $\mu=0$ and $\sigma=1$, and resize inputs to 224 × 224.
    \item \textbf{Prefer KNN over Linear Probing. Prefer KNN and Linear Probing over Fine-tuning}: Linear probing has the potential to overstate feature representation ability due to the numerous hyperparameters and ways to perform linear probing experiments. Additionally, while fine-tuning compares pretrained weights as an initialization, this tends to not be the purest indicator for representation ability and has been shown to underperform for out-of-distribution downstream tasks~\cite{kumar2022fine}.
\end{enumerate}

\medskip

\bibliographystyle{plainnat}
\bibliography{refs}

\begin{thebibliography}{59}
\providecommand{\natexlab}[1]{#1}
\providecommand{\url}[1]{\texttt{#1}}
\expandafter\ifx\csname urlstyle\endcsname\relax
  \providecommand{\doi}[1]{doi: #1}\else
  \providecommand{\doi}{doi: \begingroup \urlstyle{rm}\Url}\fi

\bibitem[Ahlswede et~al.(2023)Ahlswede, Schulz, Gava, Helber, Bischke,
  F{\"o}rster, Arias, Hees, Demir, and Kleinschmit]{ahlswede2023treesatai}
Steve Ahlswede, Christian Schulz, Christiano Gava, Patrick Helber, Benjamin
  Bischke, Michael F{\"o}rster, Florencia Arias, J{\"o}rn Hees, Beg{\"u}m
  Demir, and Birgit Kleinschmit.
\newblock Treesatai benchmark archive: A multi-sensor, multi-label dataset for
  tree species classification in remote sensing.
\newblock \emph{Earth System Science Data}, 15\penalty0 (2):\penalty0 681--695,
  2023.

\bibitem[Ayush et~al.(2021)Ayush, Uzkent, Meng, Tanmay, Burke, Lobell, and
  Ermon]{ayush2021geography}
Kumar Ayush, Burak Uzkent, Chenlin Meng, Kumar Tanmay, Marshall Burke, David
  Lobell, and Stefano Ermon.
\newblock Geography-aware self-supervised learning.
\newblock In \emph{Proceedings of the IEEE/CVF International Conference on
  Computer Vision}, pages 10181--10190, 2021.

\bibitem[Basu et~al.(2015)Basu, Ganguly, Mukhopadhyay, DiBiano, Karki, and
  Nemani]{basu2015deepsat}
Saikat Basu, Sangram Ganguly, Supratik Mukhopadhyay, Robert DiBiano, Manohar
  Karki, and Ramakrishna Nemani.
\newblock Deepsat: a learning framework for satellite imagery.
\newblock In \emph{Proceedings of the 23rd SIGSPATIAL international conference
  on advances in geographic information systems}, pages 1--10, 2015.

\bibitem[Bello et~al.(2021)Bello, Fedus, Du, Cubuk, Srinivas, Lin, Shlens, and
  Zoph]{bello2021revisiting}
Irwan Bello, William Fedus, Xianzhi Du, Ekin~Dogus Cubuk, Aravind Srinivas,
  Tsung-Yi Lin, Jonathon Shlens, and Barret Zoph.
\newblock Revisiting resnets: Improved training and scaling strategies.
\newblock \emph{Advances in Neural Information Processing Systems},
  34:\penalty0 22614--22627, 2021.

\bibitem[Caron et~al.(2021)Caron, Touvron, Misra, J{\'e}gou, Mairal,
  Bojanowski, and Joulin]{caron2021emerging}
Mathilde Caron, Hugo Touvron, Ishan Misra, Herv{\'e} J{\'e}gou, Julien Mairal,
  Piotr Bojanowski, and Armand Joulin.
\newblock Emerging properties in self-supervised vision transformers.
\newblock In \emph{Proceedings of the IEEE/CVF international conference on
  computer vision}, pages 9650--9660, 2021.

\bibitem[Cha et~al.(2023)Cha, Seo, and Lee]{cha2023billion}
Keumgang Cha, Junghoon Seo, and Taekyung Lee.
\newblock A billion-scale foundation model for remote sensing images.
\newblock \emph{arXiv preprint arXiv:2304.05215}, 2023.

\bibitem[Chen et~al.(2020)Chen, Fan, Girshick, and He]{chen2020improved}
Xinlei Chen, Haoqi Fan, Ross Girshick, and Kaiming He.
\newblock Improved baselines with momentum contrastive learning.
\newblock \emph{arXiv preprint arXiv:2003.04297}, 2020.

\bibitem[Cheng et~al.(2017)Cheng, Han, and Lu]{cheng2017remote}
Gong Cheng, Junwei Han, and Xiaoqiang Lu.
\newblock Remote sensing image scene classification: Benchmark and state of the
  art.
\newblock \emph{Proceedings of the IEEE}, 105\penalty0 (10):\penalty0
  1865--1883, 2017.

\bibitem[Chollet et~al.(2015)]{chollet2015keras}
Fran\c{c}ois Chollet et~al.
\newblock Keras.
\newblock \url{https://keras.io}, 2015.

\bibitem[Cong et~al.(2022)Cong, Khanna, Meng, Liu, Rozi, He, Burke, Lobell, and
  Ermon]{cong2022satmae}
Yezhen Cong, Samar Khanna, Chenlin Meng, Patrick Liu, Erik Rozi, Yutong He,
  Marshall Burke, David Lobell, and Stefano Ermon.
\newblock Satmae: Pre-training transformers for temporal and multi-spectral
  satellite imagery.
\newblock \emph{Advances in Neural Information Processing Systems},
  35:\penalty0 197--211, 2022.

\bibitem[Corley and Najafirad(2022)]{corley2022supervising}
Isaac Corley and Peyman Najafirad.
\newblock Supervising remote sensing change detection models with 3d surface
  semantics.
\newblock In \emph{2022 IEEE International Conference on Image Processing
  (ICIP)}, pages 3753--3757. IEEE, 2022.

\bibitem[Cover and Hart(1967)]{cover1967nearest}
Thomas Cover and Peter Hart.
\newblock Nearest neighbor pattern classification.
\newblock \emph{IEEE transactions on information theory}, 13\penalty0
  (1):\penalty0 21--27, 1967.

\bibitem[Deng et~al.(2009)Deng, Dong, Socher, Li, Li, and
  Fei-Fei]{deng2009imagenet}
Jia Deng, Wei Dong, Richard Socher, Li-Jia Li, Kai Li, and Li~Fei-Fei.
\newblock Imagenet: A large-scale hierarchical image database.
\newblock In \emph{2009 IEEE conference on computer vision and pattern
  recognition}, pages 248--255. Ieee, 2009.

\bibitem[Dosovitskiy et~al.(2020)Dosovitskiy, Beyer, Kolesnikov, Weissenborn,
  Zhai, Unterthiner, Dehghani, Minderer, Heigold, Gelly,
  et~al.]{dosovitskiy2020image}
Alexey Dosovitskiy, Lucas Beyer, Alexander Kolesnikov, Dirk Weissenborn,
  Xiaohua Zhai, Thomas Unterthiner, Mostafa Dehghani, Matthias Minderer, Georg
  Heigold, Sylvain Gelly, et~al.
\newblock An image is worth 16x16 words: Transformers for image recognition at
  scale.
\newblock \emph{arXiv preprint arXiv:2010.11929}, 2020.

\bibitem[Drusch et~al.(2012)Drusch, Del~Bello, Carlier, Colin, Fernandez,
  Gascon, Hoersch, Isola, Laberinti, Martimort, et~al.]{drusch2012sentinel}
Matthias Drusch, Umberto Del~Bello, S{\'e}bastien Carlier, Olivier Colin,
  Veronica Fernandez, Ferran Gascon, Bianca Hoersch, Claudia Isola, Paolo
  Laberinti, Philippe Martimort, et~al.
\newblock Sentinel-2: Esa's optical high-resolution mission for gmes
  operational services.
\newblock \emph{Remote sensing of Environment}, 120:\penalty0 25--36, 2012.

\bibitem[Du et~al.(2021)Du, Li, Cui, Qian, Li, and Bello]{du2021revisiting}
Xianzhi Du, Yeqing Li, Yin Cui, Rui Qian, Jing Li, and Irwan Bello.
\newblock Revisiting 3d resnets for video recognition.
\newblock \emph{arXiv preprint arXiv:2109.01696}, 2021.

\bibitem[(FSA)(2015)]{naip}
USDA Farm Service~Agency (FSA).
\newblock {National Agriculture Imagery Program (NAIP)}.
\newblock USDA Geospatial Data Gateway, 2015.

\bibitem[Fuller et~al.(2022{\natexlab{a}})Fuller, Millard, and
  Green]{fuller2022satvit}
Anthony Fuller, Koreen Millard, and James~R Green.
\newblock Satvit: Pretraining transformers for earth observation.
\newblock \emph{IEEE Geoscience and Remote Sensing Letters}, 19:\penalty0 1--5,
  2022{\natexlab{a}}.

\bibitem[Fuller et~al.(2022{\natexlab{b}})Fuller, Millard, and
  Green]{fuller2022transfer}
Anthony Fuller, Koreen Millard, and James~R Green.
\newblock Transfer learning with pretrained remote sensing transformers.
\newblock \emph{arXiv preprint arXiv:2209.14969}, 2022{\natexlab{b}}.

\bibitem[Gao et~al.(2022)Gao, Ma, Li, Dai, and Qiao]{gao2022convmae}
Peng Gao, Teli Ma, Hongsheng Li, Jifeng Dai, and Yu~Qiao.
\newblock Convmae: Masked convolution meets masked autoencoders.
\newblock \emph{arXiv preprint arXiv:2205.03892}, 2022.

\bibitem[Goyal et~al.(2021)Goyal, Law, Liu, Newell, and
  Deng]{goyal2021revisiting}
Ankit Goyal, Hei Law, Bowei Liu, Alejandro Newell, and Jia Deng.
\newblock Revisiting point cloud shape classification with a simple and
  effective baseline.
\newblock In \emph{International Conference on Machine Learning}, pages
  3809--3820. PMLR, 2021.

\bibitem[He et~al.(2016)He, Zhang, Ren, and Sun]{he2016deep}
Kaiming He, Xiangyu Zhang, Shaoqing Ren, and Jian Sun.
\newblock Deep residual learning for image recognition.
\newblock In \emph{Proceedings of the IEEE conference on computer vision and
  pattern recognition}, pages 770--778, 2016.

\bibitem[He et~al.(2020)He, Fan, Wu, Xie, and Girshick]{he2020momentum}
Kaiming He, Haoqi Fan, Yuxin Wu, Saining Xie, and Ross Girshick.
\newblock Momentum contrast for unsupervised visual representation learning.
\newblock In \emph{Proceedings of the IEEE/CVF conference on computer vision
  and pattern recognition}, pages 9729--9738, 2020.

\bibitem[He et~al.(2022)He, Chen, Xie, Li, Doll{\'a}r, and
  Girshick]{he2022masked}
Kaiming He, Xinlei Chen, Saining Xie, Yanghao Li, Piotr Doll{\'a}r, and Ross
  Girshick.
\newblock Masked autoencoders are scalable vision learners.
\newblock In \emph{Proceedings of the IEEE/CVF Conference on Computer Vision
  and Pattern Recognition}, pages 16000--16009, 2022.

\bibitem[Heidler et~al.(2023)Heidler, Mou, Hu, Jin, Li, Gan, Wen, and
  Zhu]{heidler2023self}
Konrad Heidler, Lichao Mou, Di~Hu, Pu~Jin, Guangyao Li, Chuang Gan, Ji-Rong
  Wen, and Xiao~Xiang Zhu.
\newblock Self-supervised audiovisual representation learning for remote
  sensing data.
\newblock \emph{International Journal of Applied Earth Observation and
  Geoinformation}, 116:\penalty0 103130, 2023.

\bibitem[Helber et~al.(2019)Helber, Bischke, Dengel, and
  Borth]{helber2019eurosat}
Patrick Helber, Benjamin Bischke, Andreas Dengel, and Damian Borth.
\newblock Eurosat: A novel dataset and deep learning benchmark for land use and
  land cover classification.
\newblock \emph{IEEE Journal of Selected Topics in Applied Earth Observations
  and Remote Sensing}, 12\penalty0 (7):\penalty0 2217--2226, 2019.

\bibitem[Huh et~al.(2016)Huh, Agrawal, and Efros]{huh2016makes}
Minyoung Huh, Pulkit Agrawal, and Alexei~A Efros.
\newblock What makes imagenet good for transfer learning?
\newblock \emph{arXiv preprint arXiv:1608.08614}, 2016.

\bibitem[Ilharco et~al.(2021)Ilharco, Wortsman, Wightman, Gordon, Carlini,
  Taori, Dave, Shankar, Namkoong, Miller, Hajishirzi, Farhadi, and
  Schmidt]{ilharco_gabriel_2021_5143773}
Gabriel Ilharco, Mitchell Wortsman, Ross Wightman, Cade Gordon, Nicholas
  Carlini, Rohan Taori, Achal Dave, Vaishaal Shankar, Hongseok Namkoong, John
  Miller, Hannaneh Hajishirzi, Ali Farhadi, and Ludwig Schmidt.
\newblock Openclip, July 2021.
\newblock URL \url{https://doi.org/10.5281/zenodo.5143773}.
\newblock If you use this software, please cite it as below.

\bibitem[Ke et~al.(2017)Ke, Meng, Finley, Wang, Chen, Ma, Ye, and
  Liu]{ke2017lightgbm}
Guolin Ke, Qi~Meng, Thomas Finley, Taifeng Wang, Wei Chen, Weidong Ma, Qiwei
  Ye, and Tie-Yan Liu.
\newblock Lightgbm: A highly efficient gradient boosting decision tree.
\newblock \emph{Advances in neural information processing systems}, 30, 2017.

\bibitem[Kumar et~al.(2022)Kumar, Raghunathan, Jones, Ma, and
  Liang]{kumar2022fine}
Ananya Kumar, Aditi Raghunathan, Robbie Jones, Tengyu Ma, and Percy Liang.
\newblock Fine-tuning can distort pretrained features and underperform
  out-of-distribution.
\newblock \emph{arXiv preprint arXiv:2202.10054}, 2022.

\bibitem[Lacoste et~al.(2021)Lacoste, Sherwin, Kerner, Alemohammad,
  L{\"u}tjens, Irvin, Dao, Chang, Gunturkun, Drouin, et~al.]{lacoste2021toward}
Alexandre Lacoste, Evan~David Sherwin, Hannah Kerner, Hamed Alemohammad,
  Bj{\"o}rn L{\"u}tjens, Jeremy Irvin, David Dao, Alex Chang, Mehmet Gunturkun,
  Alexandre Drouin, et~al.
\newblock Toward foundation models for earth monitoring: Proposal for a climate
  change benchmark.
\newblock \emph{arXiv preprint arXiv:2112.00570}, 2021.

\bibitem[Liu et~al.(2020)Liu, Basu, Ganguly, Mukhopadhyay, DiBiano, Karki, and
  Nemani]{liu2020deepsat}
Qun Liu, Saikat Basu, Sangram Ganguly, Supratik Mukhopadhyay, Robert DiBiano,
  Manohar Karki, and Ramakrishna Nemani.
\newblock Deepsat v2: feature augmented convolutional neural nets for satellite
  image classification.
\newblock \emph{Remote Sensing Letters}, 11\penalty0 (2):\penalty0 156--165,
  2020.

\bibitem[Mai et~al.(2023)Mai, Huang, Sun, Song, Mishra, Liu, Gao, Liu, Cong,
  Hu, et~al.]{mai2023opportunities}
Gengchen Mai, Weiming Huang, Jin Sun, Suhang Song, Deepak Mishra, Ninghao Liu,
  Song Gao, Tianming Liu, Gao Cong, Yingjie Hu, et~al.
\newblock On the opportunities and challenges of foundation models for
  geospatial artificial intelligence.
\newblock \emph{arXiv preprint arXiv:2304.06798}, 2023.

\bibitem[maintainers and contributors(2016)]{torchvision2016}
TorchVision maintainers and contributors.
\newblock Torchvision: Pytorch's computer vision library.
\newblock \url{https://github.com/pytorch/vision}, 2016.

\bibitem[Manas et~al.(2021)Manas, Lacoste, Gir{\'o}-i Nieto, Vazquez, and
  Rodriguez]{manas2021seasonal}
Oscar Manas, Alexandre Lacoste, Xavier Gir{\'o}-i Nieto, David Vazquez, and Pau
  Rodriguez.
\newblock Seasonal contrast: Unsupervised pre-training from uncurated remote
  sensing data.
\newblock In \emph{Proceedings of the IEEE/CVF International Conference on
  Computer Vision}, pages 9414--9423, 2021.

\bibitem[Mikriukov et~al.(2022)Mikriukov, Ravanbakhsh, and
  Demir]{mikriukov2022deep}
Georgii Mikriukov, Mahdyar Ravanbakhsh, and Beg{\"u}m Demir.
\newblock Deep unsupervised contrastive hashing for large-scale cross-modal
  text-image retrieval in remote sensing.
\newblock \emph{arXiv preprint arXiv:2201.08125}, 2022.

\bibitem[Musgrave et~al.(2020)Musgrave, Belongie, and Lim]{musgrave2020metric}
Kevin Musgrave, Serge Belongie, and Ser-Nam Lim.
\newblock A metric learning reality check.
\newblock In \emph{Computer Vision--ECCV 2020: 16th European Conference,
  Glasgow, UK, August 23--28, 2020, Proceedings, Part XXV 16}, pages 681--699.
  Springer, 2020.

\bibitem[Neumann et~al.(2019)Neumann, Pinto, Zhai, and
  Houlsby]{neumann2019domain}
Maxim Neumann, Andre~Susano Pinto, Xiaohua Zhai, and Neil Houlsby.
\newblock In-domain representation learning for remote sensing.
\newblock \emph{arXiv preprint arXiv:1911.06721}, 2019.

\bibitem[Neumann et~al.(2020)Neumann, Pinto, Zhai, and
  Houlsby]{neumann2020training}
Maxim Neumann, Andr{\'e}~Susano Pinto, Xiaohua Zhai, and Neil Houlsby.
\newblock Training general representations for remote sensing using in-domain
  knowledge.
\newblock In \emph{IGARSS 2020-2020 IEEE International Geoscience and Remote
  Sensing Symposium}, pages 6730--6733. IEEE, 2020.

\bibitem[Nguyen et~al.(2023)Nguyen, Brandstetter, Kapoor, Gupta, and
  Grover]{nguyen2023climax}
Tung Nguyen, Johannes Brandstetter, Ashish Kapoor, Jayesh~K Gupta, and Aditya
  Grover.
\newblock Climax: A foundation model for weather and climate.
\newblock \emph{arXiv preprint arXiv:2301.10343}, 2023.

\bibitem[Paszke et~al.(2019)Paszke, Gross, Massa, Lerer, Bradbury, Chanan,
  Killeen, Lin, Gimelshein, Antiga, Desmaison, Kopf, Yang, DeVito, Raison,
  Tejani, Chilamkurthy, Steiner, Fang, Bai, and
  Chintala]{Paszke_PyTorch_An_Imperative_2019}
Adam Paszke, Sam Gross, Francisco Massa, Adam Lerer, James Bradbury, Gregory
  Chanan, Trevor Killeen, Zeming Lin, Natalia Gimelshein, Luca Antiga, Alban
  Desmaison, Andreas Kopf, Edward Yang, Zachary DeVito, Martin Raison, Alykhan
  Tejani, Sasank Chilamkurthy, Benoit Steiner, Lu~Fang, Junjie Bai, and Soumith
  Chintala.
\newblock {PyTorch: An Imperative Style, High-Performance Deep Learning
  Library}.
\newblock In H.~Wallach, H.~Larochelle, A.~Beygelzimer, F.~d'Alché Buc,
  E.~Fox, and R.~Garnett, editors, \emph{Advances in Neural Information
  Processing Systems 32}, pages 8024--8035. Curran Associates, Inc., 2019.
\newblock URL
  \url{http://papers.neurips.cc/paper/9015-pytorch-an-imperative-style-high-performance-deep-learning-library.pdf}.

\bibitem[Reed et~al.(2022)Reed, Gupta, Li, Brockman, Funk, Clipp, Candido,
  Uyttendaele, and Darrell]{reed2022scale}
Colorado~J Reed, Ritwik Gupta, Shufan Li, Sarah Brockman, Christopher Funk,
  Brian Clipp, Salvatore Candido, Matt Uyttendaele, and Trevor Darrell.
\newblock Scale-mae: A scale-aware masked autoencoder for multiscale geospatial
  representation learning.
\newblock \emph{arXiv preprint arXiv:2212.14532}, 2022.

\bibitem[Richter et~al.(2021)Richter, Byttner, Krumnack, Wiedenroth, Schallner,
  and Shenk]{richter2021input}
Mats~L Richter, Wolf Byttner, Ulf Krumnack, Anna Wiedenroth, Ludwig Schallner,
  and Justin Shenk.
\newblock (input) size matters for cnn classifiers.
\newblock In \emph{Artificial Neural Networks and Machine Learning--ICANN 2021:
  30th International Conference on Artificial Neural Networks, Bratislava,
  Slovakia, September 14--17, 2021, Proceedings, Part II 30}, pages 133--144.
  Springer, 2021.

\bibitem[Rolf et~al.(2021)Rolf, Proctor, Carleton, Bolliger, Shankar, Ishihara,
  Recht, and Hsiang]{rolf2021generalizable}
Esther Rolf, Jonathan Proctor, Tamma Carleton, Ian Bolliger, Vaishaal Shankar,
  Miyabi Ishihara, Benjamin Recht, and Solomon Hsiang.
\newblock A generalizable and accessible approach to machine learning with
  global satellite imagery.
\newblock \emph{Nature communications}, 12\penalty0 (1):\penalty0 4392, 2021.

\bibitem[Roy et~al.(2014)Roy, Wulder, Loveland, Woodcock, Allen, Anderson,
  Helder, Irons, Johnson, Kennedy, et~al.]{roy2014landsat}
David~P Roy, Michael~A Wulder, Thomas~R Loveland, Curtis~E Woodcock, Richard~G
  Allen, Martha~C Anderson, Dennis Helder, James~R Irons, David~M Johnson,
  Robert Kennedy, et~al.
\newblock Landsat-8: Science and product vision for terrestrial global change
  research.
\newblock \emph{Remote sensing of Environment}, 145:\penalty0 154--172, 2014.

\bibitem[Stewart et~al.(2022)Stewart, Robinson, Corley, Ortiz, Ferres, and
  Banerjee]{stewart2022torchgeo}
Adam~J Stewart, Caleb Robinson, Isaac~A Corley, Anthony Ortiz, Juan M~Lavista
  Ferres, and Arindam Banerjee.
\newblock Torchgeo: deep learning with geospatial data.
\newblock In \emph{Proceedings of the 30th International Conference on Advances
  in Geographic Information Systems}, pages 1--12, 2022.

\bibitem[Sumbul et~al.(2019)Sumbul, Charfuelan, Demir, and
  Markl]{sumbul2019bigearthnet}
Gencer Sumbul, Marcela Charfuelan, Beg{\"u}m Demir, and Volker Markl.
\newblock Bigearthnet: A large-scale benchmark archive for remote sensing image
  understanding.
\newblock In \emph{IGARSS 2019-2019 IEEE International Geoscience and Remote
  Sensing Symposium}, pages 5901--5904. IEEE, 2019.

\bibitem[Sumbul et~al.(2021)Sumbul, De~Wall, Kreuziger, Marcelino, Costa,
  Benevides, Caetano, Demir, and Markl]{sumbul2021bigearthnet}
Gencer Sumbul, Arne De~Wall, Tristan Kreuziger, Filipe Marcelino, Hugo Costa,
  Pedro Benevides, Mario Caetano, Beg{\"u}m Demir, and Volker Markl.
\newblock Bigearthnet-mm: A large-scale, multimodal, multilabel benchmark
  archive for remote sensing image classification and retrieval [software and
  data sets].
\newblock \emph{IEEE Geoscience and Remote Sensing Magazine}, 9\penalty0
  (3):\penalty0 174--180, 2021.

\bibitem[Sun et~al.(2022)Sun, Wang, Lu, Zhu, Lu, He, Li, Rong, Yang, Chang,
  et~al.]{sun2022ringmo}
Xian Sun, Peijin Wang, Wanxuan Lu, Zicong Zhu, Xiaonan Lu, Qibin He, Junxi Li,
  Xuee Rong, Zhujun Yang, Hao Chang, et~al.
\newblock Ringmo: A remote sensing foundation model with masked image modeling.
\newblock \emph{IEEE Transactions on Geoscience and Remote Sensing}, 2022.

\bibitem[Torres et~al.(2012)Torres, Snoeij, Geudtner, Bibby, Davidson, Attema,
  Potin, Rommen, Floury, Brown, et~al.]{torres2012gmes}
Ramon Torres, Paul Snoeij, Dirk Geudtner, David Bibby, Malcolm Davidson, Evert
  Attema, Pierre Potin, Bj{\"O}rn Rommen, Nicolas Floury, Mike Brown, et~al.
\newblock Gmes sentinel-1 mission.
\newblock \emph{Remote sensing of environment}, 120:\penalty0 9--24, 2012.

\bibitem[Touvron et~al.(2019)Touvron, Vedaldi, Douze, and
  J{\'e}gou]{touvron2019fixing}
Hugo Touvron, Andrea Vedaldi, Matthijs Douze, and Herv{\'e} J{\'e}gou.
\newblock Fixing the train-test resolution discrepancy.
\newblock \emph{Advances in neural information processing systems}, 32, 2019.

\bibitem[Tran et~al.(2015)Tran, Bourdev, Fergus, Torresani, and
  Paluri]{tran2015learning}
Du~Tran, Lubomir Bourdev, Rob Fergus, Lorenzo Torresani, and Manohar Paluri.
\newblock Learning spatiotemporal features with 3d convolutional networks.
\newblock In \emph{Proceedings of the IEEE international conference on computer
  vision}, pages 4489--4497, 2015.

\bibitem[Tseng et~al.(2023)Tseng, Zvonkov, Purohit, Rolnick, and
  Kerner]{tseng2023lightweight}
Gabriel Tseng, Ivan Zvonkov, Mirali Purohit, David Rolnick, and Hannah Kerner.
\newblock Lightweight, pre-trained transformers for remote sensing timeseries.
\newblock \emph{arXiv preprint arXiv:2304.14065}, 2023.

\bibitem[Van~der Maaten and Hinton(2008)]{van2008visualizing}
Laurens Van~der Maaten and Geoffrey Hinton.
\newblock Visualizing data using t-sne.
\newblock \emph{Journal of machine learning research}, 9\penalty0 (11), 2008.

\bibitem[Wang et~al.(2022{\natexlab{a}})Wang, Zhang, Du, Xia, and
  Tao]{wang2022empirical}
Di~Wang, Jing Zhang, Bo~Du, Gui-Song Xia, and Dacheng Tao.
\newblock An empirical study of remote sensing pretraining.
\newblock \emph{IEEE Transactions on Geoscience and Remote Sensing},
  2022{\natexlab{a}}.

\bibitem[Wang et~al.(2022{\natexlab{b}})Wang, Braham, Xiong, Liu, Albrecht, and
  Zhu]{wang2022ssl4eo}
Yi~Wang, Nassim Ait~Ali Braham, Zhitong Xiong, Chenying Liu, Conrad~M Albrecht,
  and Xiao~Xiang Zhu.
\newblock Ssl4eo-s12: A large-scale multi-modal, multi-temporal dataset for
  self-supervised learning in earth observation.
\newblock \emph{arXiv preprint arXiv:2211.07044}, 2022{\natexlab{b}}.

\bibitem[Wightman et~al.(2023)Wightman, Raw, Soare, Arora, Ha, Reich, Guan,
  Kaczmarzyk, mrT23, Mike, SeeFun, contrastive, Rizin, Kim, Kertész, Mehta,
  Cucurull, Singh, hankyul, Tatsunami, Lavin, Zhuang, Hollemans, Rashad,
  Sameni, Shults, Lucain, Wang, Kwon, and Uchida]{ross_wightman_2023_7618837}
Ross Wightman, Nathan Raw, Alexander Soare, Aman Arora, Chris Ha, Christoph
  Reich, Fredo Guan, Jakub Kaczmarzyk, mrT23, Mike, SeeFun, contrastive,
  Mohammed Rizin, Hyeongchan Kim, Csaba Kertész, Dushyant Mehta, Guillem
  Cucurull, Kushajveer Singh, hankyul, Yuki Tatsunami, Andrew Lavin, Juntang
  Zhuang, Matthijs Hollemans, Mohamed Rashad, Sepehr Sameni, Vyacheslav Shults,
  Lucain, Xiao Wang, Yonghye Kwon, and Yusuke Uchida.
\newblock {rwightman/pytorch-image-models: v0.8.10dev0 Release}, February 2023.
\newblock URL \url{https://doi.org/10.5281/zenodo.7618837}.

\bibitem[Yang and Newsam(2010)]{yang2010bag}
Yi~Yang and Shawn Newsam.
\newblock Bag-of-visual-words and spatial extensions for land-use
  classification.
\newblock In \emph{Proceedings of the 18th SIGSPATIAL international conference
  on advances in geographic information systems}, pages 270--279, 2010.

\bibitem[Zhu et~al.(2020)Zhu, Hu, Qiu, Shi, Kang, Mou, Bagheri, Haberle, Hua,
  Huang, et~al.]{zhu2020so2sat}
Xiao~Xiang Zhu, Jingliang Hu, Chunping Qiu, Yilei Shi, Jian Kang, Lichao Mou,
  Hossein Bagheri, Matthias Haberle, Yuansheng Hua, Rong Huang, et~al.
\newblock So2sat lcz42: A benchmark data set for the classification of global
  local climate zones [software and data sets].
\newblock \emph{IEEE Geoscience and Remote Sensing Magazine}, 8\penalty0
  (3):\penalty0 76--89, 2020.

\end{thebibliography}


\newpage
\appendix

\section{Benchmarks} \label{sec:code}
To illustrate the simplicity of loading the models used in our benchmarks we show examples of how to initialize all the models used in our paper in Python on this page and an example of how to apply one of these models to a dataset on the next page. We hope that this low-overhead will encourage the use of simple baselines in future work. 

\subsection{Model Loading}
\begin{verbatim}
import timm
from torchgeo.models import (
    ResNet50_Weights,
    RCF,
    MOSAIKS,
    ImageStatistics,
    get_model
)
from torchvision.models.feature_extraction import create_feature_extractor


train_ds = ...
channels = ...
seed = ...

image_stat = ImageStatistics()
rcf = RCF(
    in_channels=channels,
    features=512,
    kernel_size=3,
    seed=seed
)
mosaiks = MOSAIKS(
    dataset=train_ds,
    in_channels=channels,
    features=512,
    kernel_size=3,
    seed=seed
)
imagenet = timm.create_model(
    "resnet50",
    in_chans=channels,
    pretrained=True,
    num_classes=0
)
random = timm.create_model(
    "resnet50",
    in_chans=channels,
    pretrained=False,
    num_classes=0
)
moco = get_model(
    "resnet50",
    weights=ResNet50_Weights.SENTINEL2_ALL_MOCO
)
moco = create_feature_extractor(
    moco,
    return_nodes=["global_pool"]
)
\end{verbatim}

\newpage
\subsection{Model Benchmarking}
Below is an example of how to benchmark a K-Nearest Neighbors classifier on MOSAIKS features from the EuroSAT multispectral dataset.

\begin{verbatim}
from torchgeo.datamodules import EuroSATDataModule
from torchgeo.models import MOSAIKS
from sklearn.neighbors import KNeighborsClassifier
from sklearn.preprocessing import StandardScaler
from sklearn.metrics import accuracy_score

channels = 13
seed = 0

dm = EuroSATDataModule("data/eurosat/", download=True)
dm.setup(stage="fit")

model = MOSAIKS(
    dataset=dm.train_dataset,
    in_channels=channels,
    features=512,
    kernel_size=3,
    seed=seed
)
model = model.eval().cuda()

x_train, y_train = [], []
for batch in dm.train_dataloader():
    with torch.no_grad():
        x = batch["image"].cuda()
        x_train.append(model(x).cpu())
        y_train.append(batch["label"])

dm.setup(stage="test")
x_test, y_test = [], []
for batch in dm.test_dataloader():
    with torch.no_grad():
        x = batch["image"].cuda()
        x_test.append(model(x).cpu())
        y_test.append(batch["label"])

x_train = torch.cat(x_train).numpy()
x_test = torch.cat(x_test).numpy()
y_train = torch.cat(y_train).numpy()
y_test = torch.cat(y_test).numpy()

scaler = StandardScaler()
x_train = scaler.fit_transform(x_train)
x_test = scaler.transform(x_test)

knn = KNeighborsClassifier(n_neighbors=5, algorithm="brute")
knn.fit(X=x_train, y=y_train)
y_pred = knn.predict(x_test)

print(accuracy_score(y_test, y_pred)) * 100.0
# 91.0740
\end{verbatim}
\end{document}